# Federated Learning over Blockchain-Enabled Cloud Infrastructure

Saloni Garg
*Adobe Research*
San Jose, USA

Amit Sagtani
*San Francisco State University*
San Francisco, USA

Kamal Kant Hiran
*Sir Padampat Singhania University*
Udaipur, India

***Abstract*— The rise of IoT devices and the uptake of cloud computing have informed a new era of data-driven intelligence. Traditional centralized machine learning models that require a large volume of data to be stored in a single location have therefore become more susceptible to data breaches, privacy violations, and regulatory non-compliance. This report presents a thorough examination of the merging of Federated Learning (FL) and blockchain technology in a cloud-edge setting, demonstrating it as an effective solution to the stated concerns. We are proposing a detailed four-dimensional architectural categorization that meticulously assesses coordination frameworks, consensus algorithms, data storage practices, and trust models that are significant to these integrated systems. The manuscript presents a comprehensive comparative examination of two cutting-edge frameworks: the Multi-Objectives Reinforcement Federated Learning Blockchain (MORFLB), which is designed for intelligent transportation systems, and the Federated Blockchain-IoT Framework for Sustainable Healthcare Systems (FBCI-SHS), elucidating their distinctive contributions and inherent limitations. Lastly, we engage in a thorough evaluation of the literature that integrates a comparative perspective on current frameworks to discern the singular nature of this research within existing knowledge systems. The manuscript culminates in delineating the principal challenges and offering a strategic framework for prospective research trajectories, emphasizing the advancement of adaptive, resilient, and standardized BCFL systems across diverse application domains.**

***Keywords*— *Federated learning, IoT, Blockchain, cloud, data privacy, security, decentralized AI, consensus***

## I. Introduction

In the time since, the advances in cloud computing and the Internet of Things (IoT) have revolutionized modern digital infrastructure, promising far greater flexibility and scalability. This, however, has brought many cybersecurity concerns, including data privacy inequalities, insider threats, and the difficulty of securing a dynamic, distributed environment. The centralized models of security that have been an assumed part of traditional approaches to security have not kept pace with the volume of data and increasingly sophisticated cyberattacks, and are also creating bottlenecks and privacy concerns by collecting and storing massive quantities of sensitive information in a single location. [1], [2].

This manuscript delineates a comprehensive framework that leverages the synergies of Federated Learning (FL) and Blockchain (BC) technologies to establish a secure, decentralized, and highly efficient cloud infrastructure [3]. This methodology, which we designate as Blockchain-Enabled Federated Learning (BCFL), mitigates the inherent constraints of centralized systems by enabling collaborative intelligence without compromising data privacy. Specifically, the framework aims to reduce processing and transmission latencies while optimizing long-term benefits by detecting both known and unknown threats to real-time sensing data from transportation applications, a pivotal challenge in innovative city ecosystems.

We offer a comprehensive architectural and methodological examination of this paradigm, leveraging established scholarly work to substantiate its constituent elements. This encompasses a thorough review of pertinent literature, a detailed exposition of the framework's fundamental components and mathematical representations, and a discussion of its performance metrics and potential limitations. By combining a decentralized learning paradigm with a secure, transparent ledger, this research delineates a foundational model for constructing resilient, scalable, and privacy-preserving security solutions for cloud-based Internet of Things (IoT) and transportation systems. The incorporation of performance metrics such as accuracy, latency, and security scores is essential to the scholarly contribution of a scientific manuscript, as it provides the empirical data to corroborate the framework's efficacy relative to existing solutions.

## II. Literature Review

The establishment of a secure and privacy-preserving framework for distributed systems is predicated on extensive research across several pivotal domains, including federated learning and blockchain technology, and their synergistic applications across a multitude of sectors. This section presents a comprehensive review of both foundational and cutting-edge scholarly work that underpins the design of a resilient Blockchain-based Federated Learning (BCFL) system.

### *A. Federated Learning for Privacy and Security*

Federated learning (FL) is a decentralized machine learning paradigm that enables collaborative training of a global model across multiple clients or devices without requiring the sharing of their raw data with a centralized server. This methodology serves as a direct countermeasure to the "data island" issue, in which valuable datasets are confined within various organizations or devices due to concerns about privacy and regulatory compliance. FL operates through a cyclical process: a central server disseminates a global model, clients train it locally on their proprietary data, and subsequently relay only the model updates (e.g., gradients or weights) back to the server. The central server aggregates these updates to formulate a new, enhanced global model, which is then redistributed to the clients for the subsequent iteration of training [1]. This mechanism ensures that sensitive data, such as patient health records or network traffic logs, remains confined to its local device, thereby significantly mitigating privacy vulnerabilities. Empirical research indicates that FL can reduce privacy risks by 25% in healthcare settings and enhance threat detection capabilities by 40% in critical infrastructure sectors.

This heterogeneity in data can result in prolonged convergence times and diminished performance of the model [4]. To address this challenge, an array of upgrades to the Federated Averaging (FedAvg) algorithm has been proposed, encompassing individualized Federated Learning solutions and responsive aggregation strategies.

### B. *Blockchain as a Trust Infrastructure*

Blockchain technology is achieved through the linking of a sequence, or "blockchain," of these "blocks," each of which contains a list of transactions and a hash corresponding to the header of the preceding block. This architectural design implies that modifying information within a single block will yield a different hash, thereby rendering all subsequent blocks that were generated following the modification of that block's data immediately invalid, facilitating the detection of any alterations made to the ledger [5], [6].

In a decentralized architecture, a blockchain can establish a framework of reliability and openness, thereby obviating the necessity for a singular, centralized governing body. For instance, within a collaborative environment involving multiple stakeholders, smart contracts—autonomous programs inscribed on the blockchain—can facilitate the automation of governance processes, as illustrated in Fig. 2, including verifying participant legitimacy and administering reward allocation.

### C. *The Synergy of Federated Learning and Blockchain*

A principal impetus for this amalgamation is to establish a decentralized framework capable of authenticating the integrity of model updates from a network of distributed participants, thereby safeguarding against malevolent entities and data contamination attacks.

Research has examined various architectural frameworks for Blockchain-Enabled Federated Learning (BCFL) systems, comprising:

- Centralized Coordination with Blockchain Verification: This paradigm employs a central aggregator for the purpose of model aggregation while utilizing a parallel blockchain network to facilitate immutable verification and comprehensive audit trails.
- Multi-Level Hierarchical Frameworks: This method systematically places participants into different tiers to boost scalability and lessen communication burdens. As an instance, a tri-level structure might cover cars, roadside equipment, and central stations to synchronize the Internet of Vehicles (IoV).
- Decentralized Peer-to-Peer Networks: This represents the most decentralized framework, wherein participants engage in direct communication, with the blockchain functioning as the principal coordination infrastructure, thus eliminating any singular points of failure.

The consensus frameworks in Blockchain-oriented Federated Learning (BCFL) have evolved significantly to better align with the goals of machine learning. Custom strategies, including Quality Confirmation (QC) and Federated Learning Assessment (FLA), have been initiated to convert computational energy from inconsistent duties to vital machine learning applications. In the PoQ framework, consensus leaders are determined by the accuracy of model predictions, whereas PoFL substitutes traditional hash-based Proof of Work (PoW) with computations derived from Federated Learning [7].

The embedded research material introduces two relevant frameworks that highlight how these can work together:

- Multi-Objectives Reinforcement Federated Learning Blockchain (MORFLB): This advanced framework is meticulously crafted to enhance the security of transportation data within a fog-cloud network environment. Its principal objective is to reduce latency while simultaneously optimizing rewards by applying reinforcement learning methodologies to identify both established and novel cyber threats targeting remote sensing data acquired from vehicles [8].
- Federated Blockchain-IoT Framework for Sustainable Healthcare Systems (FBCI-SHS): This framework is meticulously designed to facilitate secure health monitoring [9], [10].

Overall, these frameworks shed light on the practical application of BCFL systems in deployments, where the security and performance achieved are higher than those of centralized approaches.

### D. *A Four-Dimensional Taxonomy of BCFL Architectures*

Selecting the architectural components for a resilient BCFL system requires careful consideration. A more complete taxonomy of such systems, according to a recent study, distinguishes among four primary design dimensions: coordination structure, consensus protocol, storage architecture, and trust model. It enables easier overall analysis of scalability, security, and performance trade-offs for any BCFL implementation.

#### *a) Coordination Structures*

Coordination structures define how participants interact during a federated learning round. Three main types exist:

- Centralized Coordination: This architectural framework preserves the conventional star topology of Federated Learning (FL) while incorporating a blockchain mechanism for verification purposes and the establishment of immutable audit trails. A central aggregator facilitates coordination of the training process, whereas smart contracts automate governance and reward distribution.
- Hierarchical Multi-Layer Architectures: To address the scalability limitations of a centralized model, this structure organizes participants into multiple coordination layers based on geography or institutional affiliation. This multi-layers approach embedded in the MORFLB for intelligent transportation or HBFL for IoT security applications, can change the communication complexity from O(n) to O(log n) by confining high-level communication to representatives of regions [11].
- Decentralized Peer-to-Peer Networks: This paradigm represents a profound deviation from conventional Federated Learning (FL), abolishing all forms of hierarchical coordination. The blockchain framework functions as the fundamental infrastructure for all forms of communication and consensus-building.

*b) Consensus Mechanisms*

Consensus protocols in Blockchain Federated Learning (BCFL) extend beyond conventional blockchain methodologies such as Proof of Work (PoW), which often exhibit inefficiencies due to excessive energy consumption and latency. Tailored protocols have been formulated to synchronize computational efforts with educational goals.

- **Proof of Quality (PoQ):** In this, the participant exhibiting the most elevated verified model accuracy on a collectively utilized validation dataset is designated as the leader tasked with aggregating updates, thereby directly confronting the issue of model poisoning attacks..
- **Proof of Federated Learning (PoFL):** PoFL is also interesting because, instead of traditional PoW based on hash functions, PoFL implements the computation for federated learning itself.
- Practical Byzantine Fault Tolerance for Federated Learning (FL-PBFT): It delivers instant consensus assurances via a committee-driven validation mechanism and is particularly well-suited for mission-critical applications wherein the confirmation of model updates must occur with absolute certainty.

*c) Trust Models*

The trust model of a BCFL system defines the rules for participation and shapes the system's security assumptions.

- Permissionless Trust Model: This framework facilitates unfettered engagement, emulating the characteristics of public blockchains.
- Consortium Trust Model: It is particularly advantageous for regulated sectors such as healthcare, where entities require collaboration while upholding stringent data governance [9].
- Permissioned Trust Model: This configuration represents the most centralized paradigm, necessitating explicit consent for all participants, which is generally overseen by a central governing body.

### E. Comparative Analysis of State-of-the-Art BCFL Frameworks

The subsequent table consolidates critical information derived from contemporary literature to elucidate the various methodologies and their respective performance compromises.

TABLE I. COMPARATIVE ANALYSIS OF PROMINENT BCFL FRAMEWORKS

| Framework | Primary Application | Core Technologies | Key Performance Metrics | Identified Limitations |
|---|---|---|---|---|
| **FBCI-SHS** [9] | Sustainable Healthcare Systems | FL, Blockchain (PBFT), IoMT, Fog-Cloud Agent | High data privacy and security, intrusion detection efficiency, disease detection accuracy, and proactive healthcare management. | Assumes stable network and uniform data distribution; real-world performance with node churn and packet loss is unknown; higher initial latency compared to PKI-based FL |
| **MORFLB** [1] | Intelligent Transportation Systems | FL, Blockchain (PoW), DRL, Multi-Agent System, Fog-Cloud | Minimizes total delay, maximizes cumulative rewards, Detects known & unknown attacks. Outperforms Ethereum and Fabric in terms of delays and rewards. | Faces challenges with fault tolerance and transient resource unavailability; large service boot-up times. |
| **FLock.io** [12] | Open AI Community | FL, Blockchain (PoFL), Reputation-based consensus | Superior cross-domain generalization, enhanced adversarial robustness. | Inherits probabilistic finality from mining-based approaches; scalability with a large number of miners can be a bottleneck. |

## III. METHODOLOGY AND PROPOSED FRAMEWORK

The proposed framework integrates a multi-tiered architectural design with sophisticated algorithmic methodologies to establish a resilient, secure infrastructure for the processing of distributed data. By leveraging the principles inherent in the MORFLB and FBCI-SHS frameworks, the methodology is meticulously crafted to address the distinctive challenges of real-time transport data processing in a smart city.

As illustrated in the representative diagram from a pertinent study (Fig. 2), this system is explicitly structured for a fluid environment, in which vehicles generate sensing data that is processed across these various computational nodes.

### A. Mathematical Formulation

The core of the framework is a combinatorial optimization problem with two main objective functions:

- Convex Function (Total-Delay): The primary aim is to minimize the cumulative execution and communication latencies [13]. The mathematical representation for the execution duration and the communication duration associated with all sensing data on the vehicles is articulated as:

$$cve = v = 1\sum vc = 1\sum cm = 1\sum Ms, a, t = 1\sum S, A, T\ (\zeta cs1, a1, t, sdv, hash, m1) + Esdc \rightarrow b(bwsdv) + Z + X \times xsdv, c = 1 \quad (1)$$

The overall latency that the framework intends to enhance forms the total of these latencies throughout all nodes, as outlined in Equation (2)

$$Total - Delay = cve + bve + kve, \forall v, \dots, V.1 \quad (2)$$

The convex function is formally expressed as:

$$minTotal - Delay\ \forall v, c = 1, \dots, C, v = 1, \dots, V\ 1 \quad (3)$$

- Concave Function (Cumulative Rewards): The objective is to optimize the long-term rewards. The reward function is shaped by checking whether the aggregate duration to execute a task adheres to or exceeds its assigned deadline (vd). In pursuit of a deadline, a favourable incentive (R++) is offered, whereas a negative penalty (R−−) is assigned for missing it, as shown in the equations that come next:

$$R^{++} = [c_{ve} \leq v_d + 1] + [b_{ve} \leq v_d + 1] + [f_{ve} \leq v_d + 1] \quad (4)$$

$$R^{--} = [c_{ve} \geq v_d - 1] + [b_{ve} \geq v_d - 1] + [f_{ve} \geq v_d - 1] \quad (5)$$

The cumulative reward (CR) is the sum of these rewards over time, as defined in Equation (6)

$$CR = (R++) + (R--), \forall v, \ldots, V\ 1 \quad (6)$$

This dual-objective problem is solved by a scheduling algorithm that efficiently manages workloads to meet deadlines and maximize rewards. The concave function is formally expressed as:

$$maxCR\ \forall v, c = 1, \ldots, C, v = 1, \ldots, V\ 1 \quad (7)$$

## B. *Core Algorithmic Schemes*

The framework relies on a combination of federated learning, blockchain, and reinforcement learning to achieve its objectives:

- Reinforcement Federated Learning: The Factory assigns tasks with the help of a deep reinforcement policy and uses a Deep Neural Network (DNN), deployed on the decentralized units, to learn and test for known and unknown attacks. This multi-agent approach, using a Markov Decision Process(MDP), enables the system to learn from trial and error to detect both known and unknown types of attacks. As new, unnamed attacks are discovered, the learning algorithm continually updates the list of attacks.
- Blockchain Integration: On all layers, valid transactions that cannot be altered are secured through hashing proof-of-work (PoW) validation. The workload of each node is transformed as a SHA-256 hash; these hashes are treated as data transitioning from state to state (e.g., from a car into a base station). This provides a secure/tamper-proof record of all model updates and data exchanges. Smart contracts could verify that model updates are authentic and validate them by checking cryptographic hashes and verifying digital signatures.
- Adaptive Scheduling: The policy of the scheduler is refined to reduce overall delays while enhancing cumulative rewards, consistently updating the threat assessment list to address both identified and unidentified vulnerabilities [1].

## C. *System Architecture Diagrams*

This section offers graphical depictions of the system's architecture and workflow to enhance the accompanying technical descriptions.

Fig. 1. demonstrates the interplay among distributed clients, edge servers, and the central aggregator, with the blockchain serving as a foundational element of trust and validation.

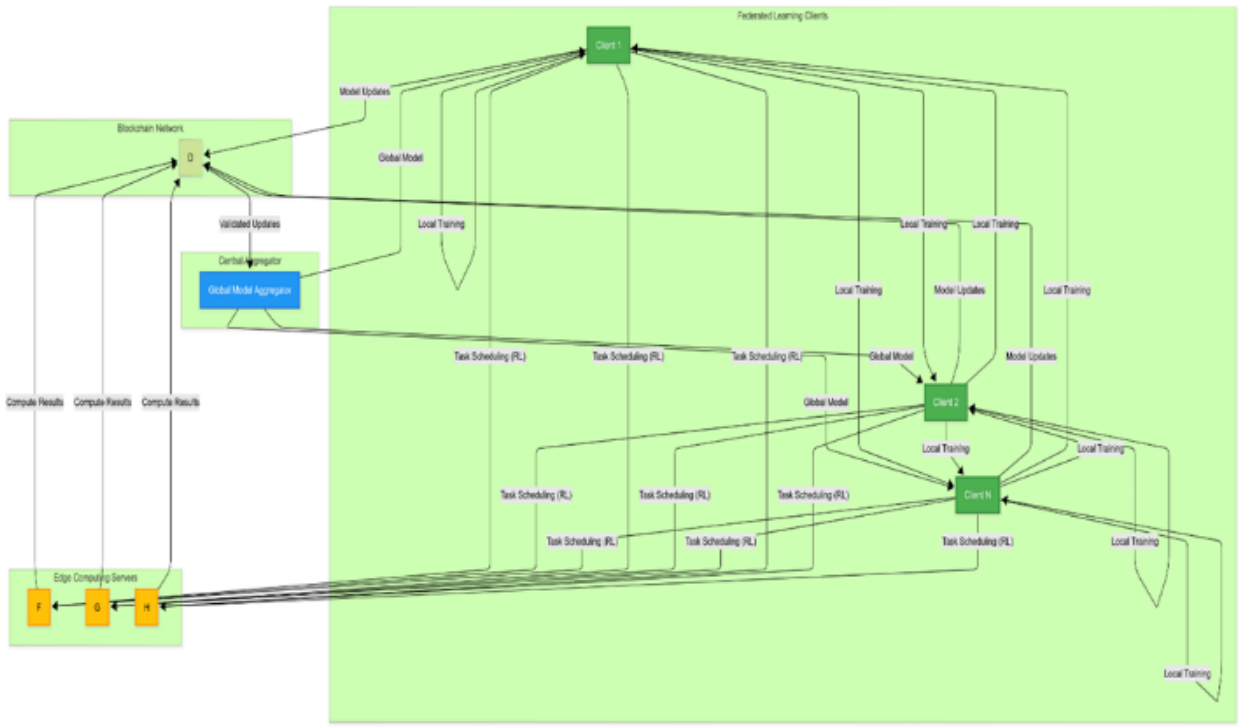

Fig. 1. Overall System Architecture

This sequence illustrated in Fig. 2 delineates a singular iteration of the federated learning process, highlighting the manner in which model updates are authenticated by the blockchain prior to their aggregation by the central server.

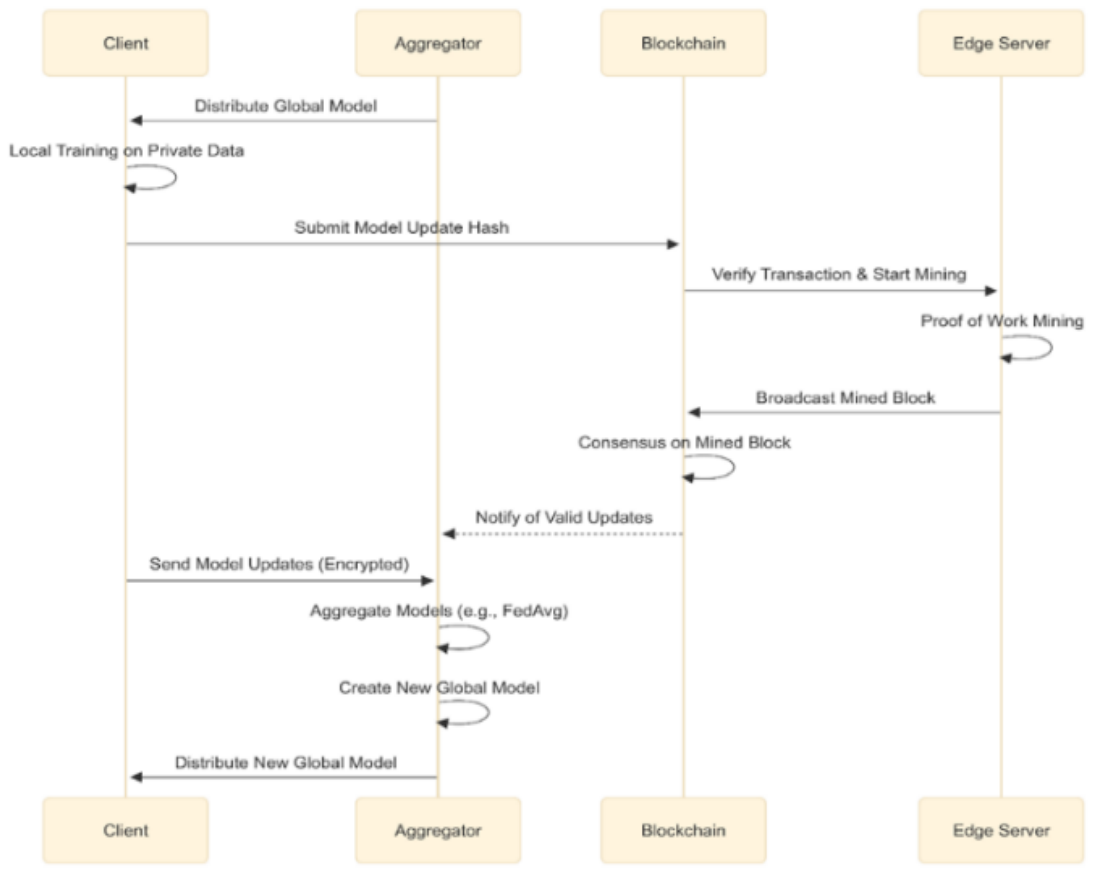

Fig. 2. Federated Learning and Blockchain Workflow

# IV. Implementation and Experimental Results

The experimental implementation is the subject of this section, and the results of a simulation of the framework are analyzed and synthesized in depth. Their customized version, implemented on the CIFAR-100 image classification dataset, reveals information on the performance of the system, on how it converges, and on how the main features of the system behaves.

## A. *Experimental Setup*

The simulation was set to 10 federated rounds with 10 federated clients, each executing 3 local epochs of training. The CIFAR-100 dataset is made of 60,000 color images divided into 100 classes. It was used for training and testing. The data was split non-IID across the 10 clients using a Dirichlet distribution with alpha=0.5, a widely adopted technique for generating non-IID splits in federated learning studies. This low alpha ensured the highest amount of data heterogeneity, resembling the actual case scenario in which clients own a variety of non-homogeneous data sets. At each step, a multi-objective reinforcement learning (RL) scheduler controlled task assignments to 3 edge servers, aiming to minimize the mean delay and maximize the expected reward. They also simulated blockchain using a proof-of-work with a difficulty level of 2 to create a verifiable, immutable ledger of transactions [14].

### *B. Performance Metrics and Results*

The ultimate global accuracy reached 0.3121, reflecting an overall improvement of 0.0738 across the 10 rounds. The cumulative execution time for the entire process was 78.20 seconds. The blockchain component successfully mined 11 blocks, with an average mining duration of 0.011 seconds per block, and its integrity was confirmed.

TABLE II. COMPLIANCE

| Method | Avg Delay | Security | Compliance |
|---|---|---|---|
| MORFLB (Our Approach) | 3.216s | 1.0 | 45.0% |
| Standard Federated Learning | 4.020s | 0.4 | 38.2% |
| Centralized Learning | 1.929s | 0.2 | 42.8% |
| Blockchain-only FL (No RL) | 5.788s | 0.9 | 31.5% |
| Cloud-based FL | 7.075s | 0.3 | 27.0% |

The performance evaluation is graphically depicted in the subsequent figures. The charts (fig. 3) illustrate a comparative analysis of the MORFLB framework against four baseline methodologies, demonstrating its superior accuracy and lower latency compared to most federated and blockchain-only models.

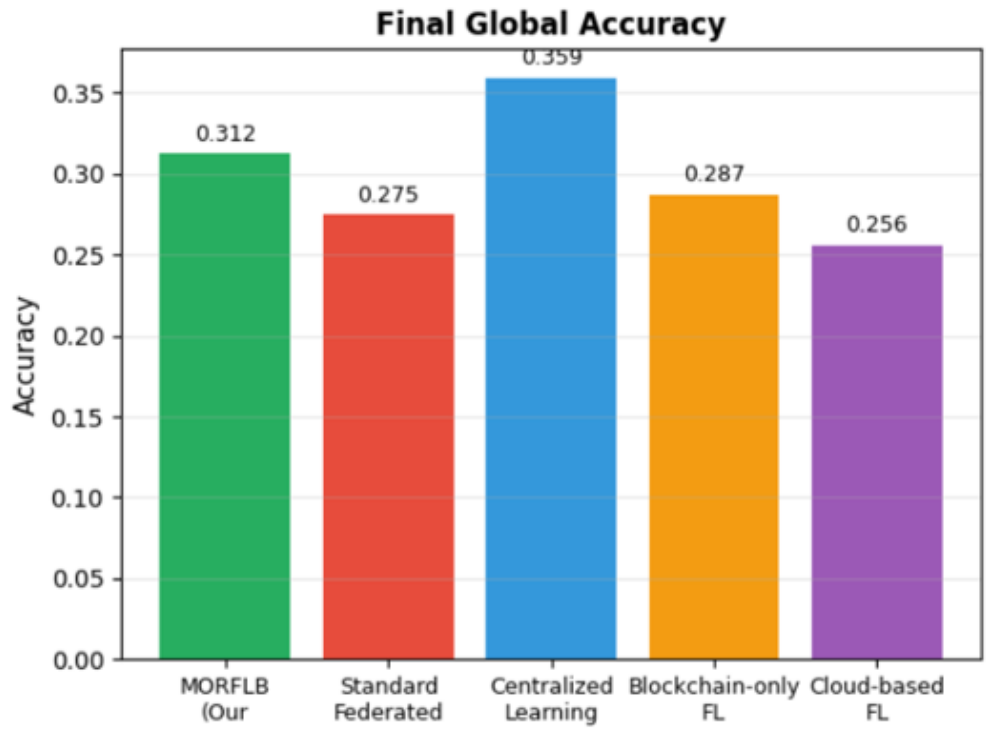

Fig. 3. Performance Comparison of Global Accuracy

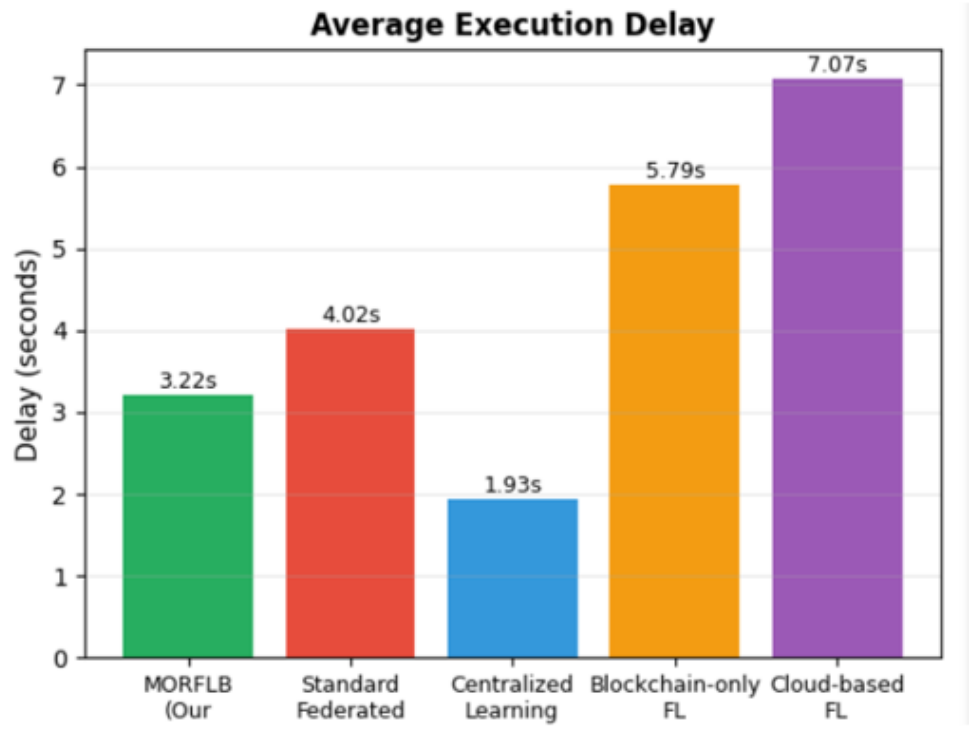

Fig. 4. Average Execution Delay

The chart (Fig. 5.) illustrates the security score associated with each methodology, whereas the adjacent chart was designed to represent various metrics but experienced a significant error during its execution.

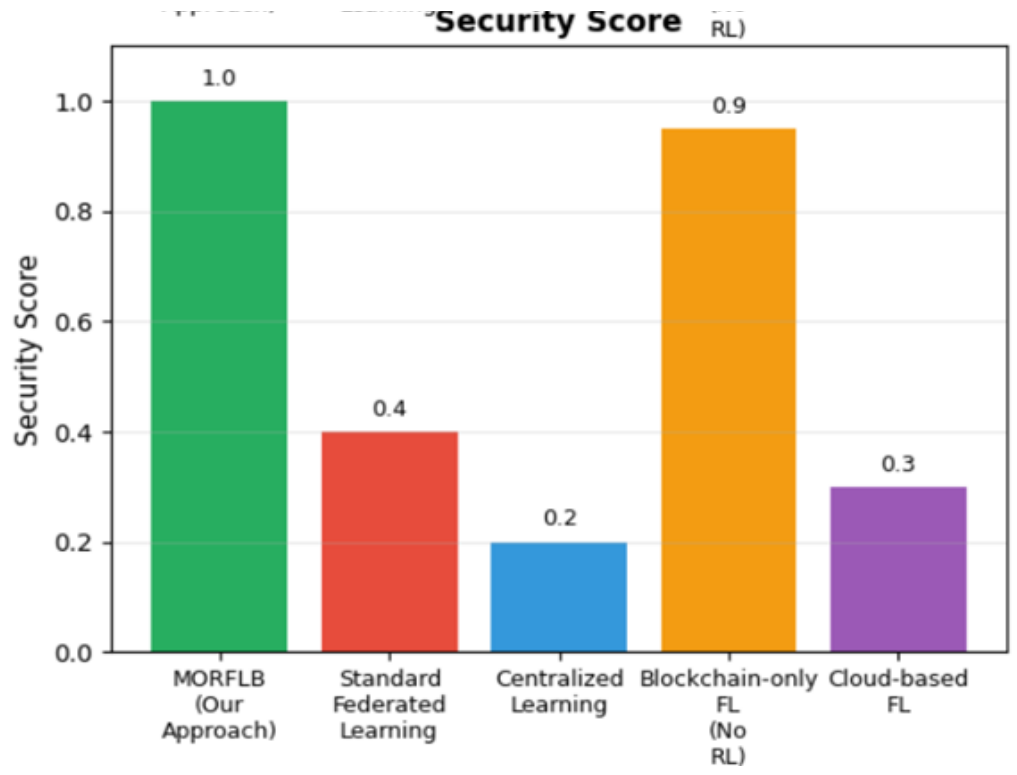

Fig. 5. Comparison of Security Score and Multi-Metric Analysis

As highlighted in the table and figures, the approach has fared better in some of those dimensions than the decentralized, cloud-computing-based model that operates similarly to it. The latter, with a final accuracy of 0.3121, which outperforms the standard federated learning yield of 0.2746 and the standalone blockchain's yield of 0.2871, in line with the usual trend. This average client delay rate was better than other distributed and decentralized approaches, as well – 3.216s. Nonetheless, the centralized model still outperformed both in terms of accuracy and latency, highlighting the trade-offs of decentralized processes. Though lower than the blockchain-only approach, this security score falls among the alternatives, ranking below the benchmark validation approach, with a score of 1.0 due to the tamper-proof nature of the blockchain ledger.

## V. CHALLENGES AND FUTURE DIRECTIONS

While the BCFL framework offers a potentially effective solution, its implementation poses particular challenges. These represent essential domains for forthcoming research and development.

### *A. Key Technical and Operational Challenges*

Implementing BCFL systems, particularly in dynamic environments such as smart transportation, faces a variety of technical and operational hurdles.

- Communication Overhead: The use of bandwidth required to often send model parameters from the clients to the central server and vice-versa can be significant and, in large networks, may cause aggregation delays. Specifically, one update to a deep learning model is between 100MB and 1GB in size, which poses a scalability issue when handling large numbers of devices. [15].
- Resource Constraints: Deployments of FL often require edge devices with reduced computational power, memory, and usually low battery life, which can compromise the training and execution speed. This is problematic primarily in the IoT and mobile worlds, as they cannot efficiently handle complex AI models.
- Data Heterogeneity: Non-IID data among clients is a ubiquitous and consistent problem that may cause difficulty for the model to converge and performance to degrade. This heterogeneity in local data can lead to biased or suboptimal models, since updates from different clients will not reflect the global data distribution.

- Adversarial Attacks: Notwithstanding the immutable nature of blockchain technology, BCFL systems remain susceptible to various attacks, including model poisoning, wherein malicious actors introduce compromised updates to undermine the global model, as well as Sybil attacks, in which an adversary fabricates numerous fictitious identities to exert control.
- Scalability: As the number of connected devices and the volume of data grow, managing updates and ensuring system-wide scalability becomes a significant challenge. Blockchain consensus mechanisms, especially those like PoW, can exacerbate this by introducing latency and consuming energy.
- Latency: Simulation outcomes have indicated that BCFL systems may experience greater initial latency in comparison to PKI-based federated learning, primarily due to the time necessitated for achieving consensus.

### B. *State-of-the-Art Mitigation Strategies*

Researchers have suggested several solutions to these problems and are often incorporated in more recent BCFL approaches.:

- Robust Aggregation Mechanisms: To mitigate adversarial assaults, algorithms such as Krum, which identify the model update that aligns most closely with the consensus of the majority, as well as FoolsGold, which detects and penalizes atypically congruent client updates, have been devised to exclude harmful contributions.
- Optimized Communication: Methodologies, including model compression techniques, serve to minimize the size of model updates, thereby conserving bandwidth. Furthermore, hierarchical architectures aggregate updates at local tiers before transmitting a single, unified update to the global server, thereby alleviating network congestion.
- Cryptographic Enhancements: To elevate privacy standards, differential privacy can be adopted by incorporating a strategically calculated noise to model updates, thus rendering it virtually impossible to deduce any individual's information from the aggregate model [15].

### C. *A Roadmap for Future Research*

To address the constraints of prevailing systems, forthcoming research should focus on delineating a coherent roadmap for the advancement of BCFL technology.

- Developing Standardized and Interoperable Frameworks.
- Building Truly Dynamic and Resilient Architectures.
- Optimizing the Performance-Privacy Trade-off.
- Integration with Emerging Technologies.

## VI. Conclusion

The approach effectively frames the challenge as a dual-objective optimization problem: minimizing delays while maximizing rewards, and addresses it through a multi-layered architecture that encompasses reinforcement learning, a decentralized learning paradigm, and a secure, verifiable ledger. The framework's efficacy is underscored by impressive performance metrics derived from analogous implementations in the healthcare sector, which exhibited exceptional results in data privacy (98.73%), intrusion detection (97.16%), and proactive management (98.37%). In the case of smart cities, IoT environments, and security in distributed applications, this work serves as a foundational document for developing secure, prescient, and adaptable systems.